\documentclass[conference]{IEEEtran} 
\usepackage[utf8]{inputenc}
\usepackage{float}
\usepackage{caption}
\usepackage{graphicx} 
\usepackage{amsmath}
\usepackage{amssymb}
\usepackage{amsfonts}
\usepackage{mathtools}
\usepackage{bm}
\usepackage{comment}
\usepackage{subfig}
\usepackage{epsfig}

\def\BibTeX{{\rm B\kern-.05em{\sc i\kern-.025em b}\kern-.08em
    T\kern-.1667em\lower.7ex\hbox{E}\kern-.125emX}}

\begin{document}   
\title{Optimal Cosserat-based deformation control for robotic manipulation of linear objects}

\author{\IEEEauthorblockN{Azad Artinian, Quentin Huet, Faïz Ben Amar, Véronique Perdereau} 
}

%\author{\IEEEauthorblockN{Artinian Azad}
%\and
%\IEEEauthorblockN{Huet Quentin}
%\and
%\IEEEauthorblockN{Ben Amar Faïz}
%\and
%\IEEEauthorblockN{Perdereau Véronique}}

\maketitle

\begin{abstract}
Deformable object manipulation is a challenging research subject that draws a growing interest in the robotic field as new methods to tackle this problem have emerged. 
So far, most of the proposed approaches in the literature focused solely on shape control. The strain applied to the object is disregarded, thus excluding a large part of industrial applications where fragile products are manipulated, like the demolding of rubber and plastic objects, or the handling of food. These applications require a tradeoff between accuracy and careful manipulation in order to  preserve the manipulated object.
In this article, we propose an approach to optimally control the deformation of linear and planar deformable objects while also minimizing the deformation energy of the object. 
First, we modify the framework initially developed for linear soft robot control in order to adapt it to deformable object robotic manipulation. To do so, we reformulate the problem as an optimization problem where the whole shape of the object is taken into account instead of solely focusing on the tip of the object's position and orientation. 
We then include an energy term in the cost function to find the solution that minimizes the potential elastic energy in the manipulated object while reaching the desired shape. 
Solutions to problems with high non-linearities are notoriously difficult to find and sensitive to local minima. 
We define intermediate optimal steps connecting the known initial and final configurations of the object and sequentially solve the problem, thus enhancing the robustness of the algorithm and ensuring the optimality of the solution. 
%Each solution is then used as input for the next step's problem, therefore enhancing the robustness of the algorithm and ensuring the optimality of the solution. 
The intermediate optimal configurations are then used to define an end-effector trajectory for the robot to deform the object from an initial to the desired configuration. 
\end{abstract}


\begingroup

\section{Introduction}

Robotic manipulation of deformable objects has a wide range of applications, from food handling \cite{wang2012finite} to cable manipulation \cite{zhu2018dual}, \cite{bretl2014quasi} in order to automate the production of motors \cite{roussel2014deformable, shah2016towards}, garment manipulation \cite{matas2018sim} or robotic surgery \cite{patil2010toward, leizea2015real}. Deformable objects are everywhere in our daily lives and the industry. However, the study of deformable objects has not received as much attention in the research community as rigid object manipulation. This is mainly due to the challenging problems manipulating these objects raise. Deformable objects have very high degrees of freedom which not only make the classic rigid object manipulation techniques impossible to apply \cite{jia2013optimal} but also very hard to model accurately. Recent advances in computer graphics, machine learning, and hardware development have resulted in new modeling and simulation tools and techniques that allow overcoming some of the classical issues raised when manipulating these objects \cite{yin2021modeling}. As a result, the demand for robotic solutions to manipulate deformable objects has emerged \cite{sanchez2018robotic}.
To manipulate these objects successfully, creating a connection between the actions performed by the robot and the behavior of the manipulated object is necessary. Among the many recent articles in the literature attempting to handle this difficult issue, two classes of methods emerge: data-driven and model-based approaches.

%Since a model of the object can be difficult to obtain, Navarro-Alarcon in a series of work \cite{navarro2013model}, \cite{navarro2014visual}, \cite{navarro2016automatic}, \cite{navarro2017fourier} relied on vision based techniques to control the shape of deformable objects. Although this method allows for a computationally efficient shape control, it is sensible to occlusions and not suited for applications where the object can fold and hide the tracked points. 

\par To address the high dimensionality of deformable objects and the modeling difficulties that derive from it, data-driven approaches have been used to solve deformable object manipulation tasks, especially garments and planar objects \cite{matas2018sim, schulman2013case}.
%like garment folding \cite{matas2018sim} or suturing \cite{schulman2013case}.
These methods are usually formulated as a supervised learning problem where the robot must replicate a desired behavior. The training data typically originate from human demonstration \cite{schulman2013case}. These approaches have the advantage of not relying on a specific model but they are often hard to generalize from a small amount of training data \cite{pignat2017learning, schulman2016learning}. Other methods use data from robotic exploration, and although an exhaustive exploration is rarely used, reinforcement learning approaches are often initialized with human demonstration \cite{tsurumine2019deep}. The improvement in simulators opened the possibility to learn politics in simulated environments \cite{matas2018sim, clegg2018learning}. However, the policies learned in simulation often face a sim-to-real gap, especially for deformable objects which are difficult to represent accurately in a simulator.

\par Model-based methods rely on a model of the object to predict the deformations. There is a wide range of models with different properties. Less physically realistic models like mass-spring systems \cite{hirai2001robust} and position-based dynamics \cite{bender2015position} tend to be computationally effective but can sometimes lack precision, especially for larger deformations for mass-spring systems while position-based dynamics does not accurately model force effect. 
Although most models are determined beforehand, approaches relying on visual information to estimate a model online have also been proposed \cite{navarro2013model, navarro2014visual, navarro2016automatic, navarro2017fourier}.

\par Recently, physically realistic continuum models such as Finite elements \cite{duenser2018interactive} have been used to model deformable objects for robotic applications despite their high computational cost. In \cite{ficuciello2018fem}, the authors, based on the framework proposed in \cite{coevoet2017software}, \cite{duriez2016framework} for soft robot real-time control, proposed an inversion method of the finite element model of a deformable object. The inverse FEM model is then used to determine in real-time the controls to apply at the fingertips of a robotic hand to deform the object towards a desired shape \cite{faure2012sofa}. 
With a similar approach, Cosserat theory has been used to successfully model, control \cite{rucker2010geometrically, rucker2011statics} and estimate \cite{lilge2022continuum} the state of linear deformable robots.
The Cosserat theory provides one of the most accurate mechanical model while still being computationally efficient. The authors in \cite{campisano2021closed} proposed a real-time closed-loop control strategy courant synonymesof a soft linear robot using the Cosserat rod model. It is especially suited to beam-like and linear objects which makes this model adequate for cables, ropes, and linear robot control \cite{pai2002strands}.
The formulation of the problem of shape control is very similar to the underactuated soft robot control problem. There is an analogy between the actuators in soft robot control and contact points in deformable object manipulations \cite{ficuciello2018fem}. 

\par In this article, we propose a method to generate optimal robot end-effector trajectories to solve constrained deformable object manipulation problems in industrial environments. We achieve this by adapting the framework originally developed for soft robot control in \cite{rucker2010geometrically, rucker2011statics, campisano2021closed} to deformable linear object manipulation and by combining the Cosserat rod model used to represent the object with an optimization approach.
We then define intermediate optimal steps in order to generate the optimal robot trajectory to solve the problem with respect to its specificities. 

%Contrary to previous work on soft robot control and deformable object manipulation, we don't focus solely on position control. Instead we define a general objective function where the weights on each term can be set according to the constraints and objectives specific to the task.
%This approach is therefore more suited to solve complex constrained manipulation tasks with multiple objectives. 

%Although a formulation of Cosserat surface exist that extend the model to planar objects \cite{altenbach2010generalized}, it is not used in practice in robotics due to the partial differential equation formulation that makes this model hardly compatible with real time applications.
%The main advantage of the cosserat rod model over FEM is accuracy, matching the cosserat performance would require a very fine FEM mesh and both models are very dependent on the estimation of viscoelastic parameters \cite{petit2017tracking}. 

\par
The paper is organized as follows. First, we introduce the general framework for the Cosserat rod model and how the constitutive equations are derived. Then we present our approach to adapt the problem from soft robot control to deformable object manipulation and how we formulate and solve the optimization problem, resulting in an end-effector trajectory to reach the desired deformed shape. Lastly, we illustrate this approach with a manipulation application where we control the shape of a linear clamped object with a robotic arm.

\section{Classical static Cosserat rod model}

\noindent To define the shape of a linear deformable object, we have to consider the rod kinematics that provides the position of the curve in $\mathbb{R}^{3}$ and its orientation. The rod is spatially discretized so that the model constitutive equations are derived from the static equilibrium at any point of the rod. 
These equations give a relation between the forces and moments applied on the rod and its position and orientation. Therefore, the position and orientation of every section of the rod are entirely defined by the external loads, initial conditions, and material properties.
In the following section we will rely on the methods developed by Antman in \cite{keener2005ss} and later applied to soft robotic manipulators in \cite{trivedi2008geometrically, jones2009three, rucker2010geometrically, rucker2011statics}.
Although these works focus on soft robotic manipulators, the geometric framework and formalism used to define the shape of a linear object can be used for robotic manipulation.

\subsection{Kinematics}
\noindent We consider a linear Cosserat rod of length L defined by the position $p(s) \in \mathbb{R}^{3}$ and orientation $R(s) \in SO(3)$ of every point of the reference parameter s $\in [0, L]$ of the rod. Therefore, $p(s_{1})$ gives the position in $\mathbb{R}^{3}$ of the point $s_{1}$ $m$ from the origin of the undeformed rod while $R(s_{1})$ is the orientation in the global frame attached to the point $s_{1}$. As presented in \cite{rucker2011statics}, we can use the homogeneous rigid-body transformation $g(s) \in SE(3)$ to describe the configuration of any point $s$ of the rod:

\begin{center}
$g(s) = \begin{bmatrix} R(s) & p(s) \\ 0 & 1 \end{bmatrix}$
\end{center}

$v(s)$ and $u(s)$ represent the local changes in translation and rotation of an infinitesimally small section of the Cosserat rod with respect to s. They indicate the evolution of $g(s)$ along s: 

\begin{equation}\label{eq:p et R}
\begin{aligned}
    \dot{p}(s) = R(s)v(s), \\
    \dot{R}(s) = R(s)\hat{u}(s)
\end{aligned}
\end{equation}

Here, the hat $\hat{}$ represents the bijective mapping from a vector to a skew-symmetric matrix while the dot indicates the spatial derivative along s.

\subsection{Static equilibrium}

\noindent Consider an arbitrary segment of the rod from a point $s_{1}$ to a point $s_{2}$ with $s_{1} < s_{2}$ and $(s_{1}, s_{2}) \in [0, L]^{2}$. The static equilibrium equation on this segment gives: 

\begin{equation}
    n(s_{2}) - n(s_{1}) + \int_{s_{1}}^{s_{2}}f(\sigma)d\sigma \ = 0, 
\label{static force}
\end{equation}

\begin{equation}
\begin{aligned}
    &m(s_{2}) + p(s_{2})\times n(s_{2}) - m(s_{1}) -p(s_{1})\times n(s_{1}) \\
    & + \int_{s_{1}}^{s_{2}}(p(\sigma)\times f(\sigma) + l(\sigma))d\sigma = 0,
\end{aligned}
\label{static_mom}
\end{equation}

Where $n(s_{2})$, $m(s_{2})$ are respectively the internal forces and moments applied by the $[s_{1}, s_{2}]$ segment to the next segment of the rod ($]s_{2}, L]$) while $n(s_{1})$, $m(s_{1})$ are respectively the internal forces and moments applied by this segment to the previous segment of the rod ($]0, s_{1}[$).
$f$ and $l$ are the force and moment distributions applied along the segment and the terms $\int_{s_{1}}^{s_{2}} f(\sigma)d\sigma $ and $\int_{s_{1}}^{s_{2}}(p(\sigma)\times f(\sigma) + l(\sigma))d\sigma$, therefore, represent the external loads applied to the whole segment as shown in Figure \ref{dessin_cosserat} for a section from $0$ to $L$.
By taking the partial derivative form with respect to s the equations \ref{static force} and \ref{static_mom} become the constitutive equations for internal force and moment: 

\begin{equation}\label{eq: n m v2}
\begin{aligned}
    &\dot{n}(s) + f(s) = 0, \\
    &\dot{m}(s) + \dot{p}(s)\times n(s) + l(s) = 0
\end{aligned}
\end{equation}

    \begin{figure}[!t]
        \centering
        \includegraphics[scale=0.69]{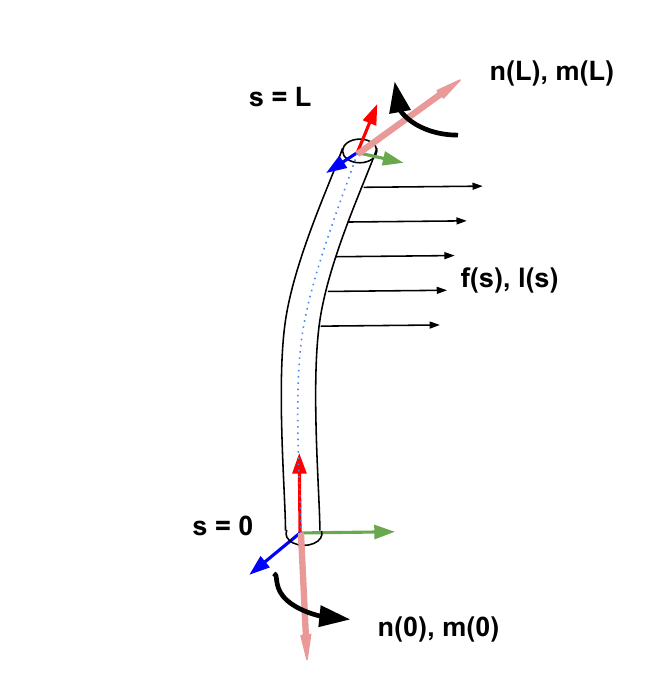}
        \caption{Section from 0 to L of a Cosserat rod, deformed under the action of external forces f and moments l. Internal forces n and moments m are also represented at s = 0 and s = L.}
        \label{dessin_cosserat}
    \end{figure}

\subsection{Constitutive Equations}
\noindent In this section, we use the formulation developed by Rucker in a series of works (\cite{rucker2010geometrically, rucker2011statics, rucker2011computing}.
The constitutive equations relate the deformation of the rod to the internal forces and moments and external loads applied on the rod. The rod is made of a linear elastic material that respects Hooke's law: $F = k\Delta x$. The notation $\sim$ refers to the undeformed configuration of the rod. Therefore $\tilde{p}(s)$, $\tilde{R}(s)$, $\tilde{v}(s)$ and $\tilde{u}(s)$ refer to the initial rest position, orientation and local changes in translation and rotation of the reference rod.

The stiffness matrices, for shear and extension $K_{tra}$ and for bending and torsion $K_{rot}$ are defined as: 

\begin{equation}
\begin{aligned}
& K_{tra}=\begin{bmatrix} A(s)G & 0 & 0 \\ 0 & A(s)G & 0 \\ 0 & 0 & A(s)E \end{bmatrix}\\
& K_{rot}=\begin{bmatrix} I_{x}(s)E & 0 & 0 \\ 0 & I_{y}(s)E & 0 \\ 0 & 0 & J(s)G \end{bmatrix}
\end{aligned}
\end{equation}

Where $I_{x}$ and $I_{y}$ are the second moments of area of the rod cross-section, $J$ is the polar moment of inertia of the cross-section ($J = I_{x} + I_{y}$). $E$ and $G$ are the Young's and shear moduli. From the linear constitutive laws are then derived the following relations between internal forces and moments and local changes $v(s)$ and $u(s)$, expressed in the local coordinate frame:

\begin{equation}\label{eq:n et m}
\begin{aligned}
    n(s) = R(s)K_{tra}(s)(v(s) - \tilde{v}(s)), \\
    m(s) = R(s)K_{rot}(s)(u(s) - \tilde{u}(s))
\end{aligned}
\end{equation}

From \ref{eq:n et m} it is possible to obtain an expression of v and u with respect to s: 

\begin{equation}\label{eq:v et u}
\begin{aligned}
    v(s) = K_{tra}^{-1}R^{T}(s)n(s) + \tilde{v}(s), \\
    u(s) = K_{rot}^{-1}R^{T}(s)m(s) + \tilde{u}(s)
\end{aligned}    
\end{equation}

Equations \ref{eq:p et R}, \ref{eq: n m v2} and \ref{eq:v et u} then constitute the following system of differential equations:

\begin{equation}\label{eq:v et u}
\begin{aligned}
    &\dot{p} = Rv,   \qquad      & v = K_{tra}^{-1}R^{T}n + \tilde{v}, \\
    &\dot{R} = R\hat{u}, \qquad  & u = K_{rot}^{-1}R^{T}m + \tilde{u}, \\
    &\dot{n} = -f, \\
    &\dot{m} = -\dot{p} \times n - l
\end{aligned}    
\end{equation}

%Another equivalent system of equations can be obtained by using the derivative of (\ref{eq:n et m}) and replacing it in (\ref{static force}) and (\ref{static mom}). $v$ and $u$ are then used as state variables instead of $n$ and $m$: 

%\begin{equation}\label{eq:v et u}
%\begin{aligned}
%    &\dot{p} = Rv, \\
%    &\dot{R} = R\hat{u}, \\
%    &\dot{v} = \dot{\tilde{v}} - K_{tra}^{-1}(\hat{u}K_{tra}\Delta v + R^{T}f) , \\
%    &\dot{u} = \dot{\tilde{u}} - K_{rot}^{-1}(\hat{u}K_{rot}\Delta u + \hat{v}K_{tra} \Delta v + R^{T}l)
%\end{aligned}    
%\end{equation}

%Where $\Delta v = v - \tilde{v}$ and $\Delta u = u - \tilde{u}$.

\section{Model-based optimization under constraints}

\subsection{Shape Control Approach}

\noindent In soft robot control applications \cite{rucker2010geometrically, rucker2011computing, campisano2021closed}, the emphasis is put on the actuation of the tip of the robot. In these works, the authors find the solution to the non-linear equations \ref{eq:v et u} that respect the desired position and orientation of the tip of the robot. 
While this approach is efficient when controlling the position of a single point and in general for soft robot actuation, it is not adapted for general linear object manipulation as it struggles to achieve complex object shapes. To accurately control the shape of a deformable object with a robotic arm, it is needed to solve an underactuated problem with limited grasping points to deform an object with a large number of degrees of freedom. 

\par
We consider the configuration of the rod defined by the configuration of every point of the rod (ie. position $p$, orientation $O$, internal forces $n$, and moments $m$ of every point of the rod).
In order to adapt the approach to deformable object manipulation, we define a low-level representation consisting of a set of $k$ objective points carefully selected such that their position represents the desired shape of the manipulated object. See Figure \ref{pcont_vs_distance} where the objective points (red) represent the desired shape. 
The objective is to deform the manipulated object in order to match the shape defined by the set of objective points. In \cite{duenser2018interactive, ficuciello2018fem}, the authors minimize the position (and orientation) error between a set of desired positions and a set of pre-defined control points on the object. This approach works best when the number of pre-defined points whose positions are controlled is equal to the number of actuated points, hence why it is used in soft robot actuation.
However, for shape control applications it is needed to control multiple points with partially coupled positions. These coupling constraints considerably limit the set of feasible solutions even though a solution that produces the desired shape exists when picking different control points with different couplings. Figure \ref{pcont_vs_distance} illustrates this issue. It is impossible for the control points (blue) to match the position of the objective points (red) because it would require the rod to both stretch and compress at different locations, which would require at least three grasping points, although a solution exists when selecting other control points. 

    \begin{figure}[!t]
        \centering
        \includegraphics[scale=0.45]{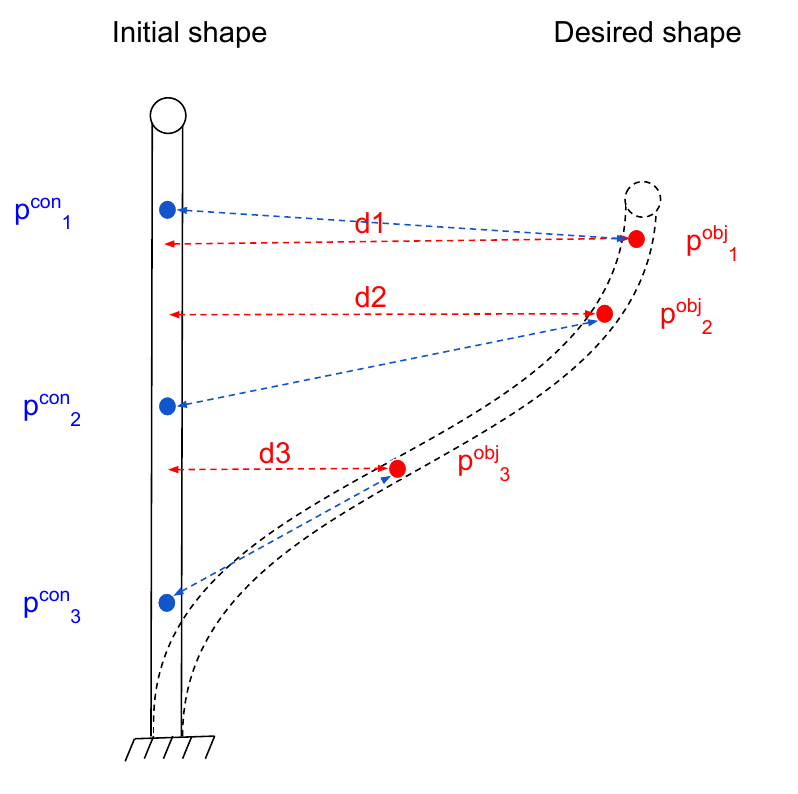}  
        \caption{Representation of the issue with the classic control points (blue) - objective points (red). We propose instead to minimize the distances $d_{1}$, $d_{2}$, $d_{3}$}
        \label{pcont_vs_distance}
    \end{figure}

We propose instead to consider the curve $\Gamma$ formed by the linear object in the 3D space and to minimize the distance error between the set of $k$ objective points and $\Gamma$. Thereby we do not control the position of a few control points and instead control the shape of the object so that the distance between $\Gamma$ and the objective positions is minimal. The result is a curve $\Gamma$ that passes through every desired point but without additional constraints on the positions of specific control points.
The number of objective points $k$ is defined arbitrarily but the higher $k$ is, the more important the constraints on the desired configuration are and the harder it is to find a solution. We then approximate the target shape with B-splines to ensure its feasibility.  

\subsection{Deformation Energy}

\noindent While this approach is more effective in finding solutions that respect a desired shape, it has the disadvantage of generating multiple solutions. %%%%%%%%%%% figure explicative %%%%%%%%%%%%%%%
To rank these solutions, we introduce the deformation energy $E$ as a secondary objective. In \cite{duenser2018interactive}, the authors compute the FEM energy to determine a stable robot pose. In our case, we search the solution that minimizes $E$, which ensures the internal strains in the object are minimal. This is especially useful for industrial applications where objects must be handled with care. 
We suppose the system is quasi-static and the gravity negligible: 

\begin{equation}\label{energy}
\begin{aligned}
    E & = \int_{0}^{L}e(s)ds, \\
      & = \int_{0}^{L}\frac{1}{2}(v(s)-\Tilde{v}(s))^{T}(s)K_{tra}(s)(v(s)-\Tilde{v}(s))\\
      &+ \frac{1}{2}(u(s)-\Tilde{u}(s))^{T}K_{rot}(s)(u(s)-\Tilde{u}(s))ds
\end{aligned}
\end{equation}

%Where $e$ is the energy in a section of the cosserat rod of length $L$.
Since integrating the Cosserat equations provides the full configuration of the rod, the only unknowns are the initial conditions that correspond to the configuration of the first point of the rod.
The solution are the initial conditions $p(0)$, $O(0)$ $n(0)$, $m(0)$ that lead to an optimal configuration of the rod. The boundary conditions depend on the type of grasping at the limits of the object, for realistic grasping situations, additional constraints must be set to reflect the grasping uncertainty (ie. $n_{x} = 0$ or $m_{x} = 0$ if the grasp does not enable forces or moments along the $x$ axis).
For a linear object of length $L$, clamped at one extremity and grasped by a robotic arm at the other, the objective function is defined as:  

\begin{equation}
\begin{aligned}
&\min_{n(0),m(0)} F  = \frac{W_{1}}{2L}\int_{0}^{L}(v(s)-\Tilde{v}(s))^{T}K_{tra}(s)\\
&(v(s)-\Tilde{v}(s))+ (u(s)-\Tilde{u}(s))^{T}K_{rot}(s)\\
&(u(s)-\Tilde{u}(s))ds + W_{2}\sum_{i=1}^{k} dist(\Gamma, p^{obj}(s_{i}),
\end{aligned}
\label{cost_fun}
\end{equation}

%& \text{subject to:}\hspace{3mm}
\hspace{2mm} subject to:
$$
\begin{cases}
    &p(s=L) \subset Ws,\\
    &\dot{p} = Rv, \\
    &\dot{R} = R\hat{u},\\
    &\dot{n} = -f, \\
    &\dot{m} = -\dot{p} \times n - l
\end{cases}
$$

\label{cost_fun_cons}

\par We have a weighted ($W_{1}, W_{2}$) objective function where $p^{obj}(s_{i = 1..k})$ are the objective points positions and $Ws$ is the operational workspace of the robot's end-effector (see Figure \ref{robot_ws}).

    \begin{figure}[htb]
        \centering
        \includegraphics[scale=0.42]{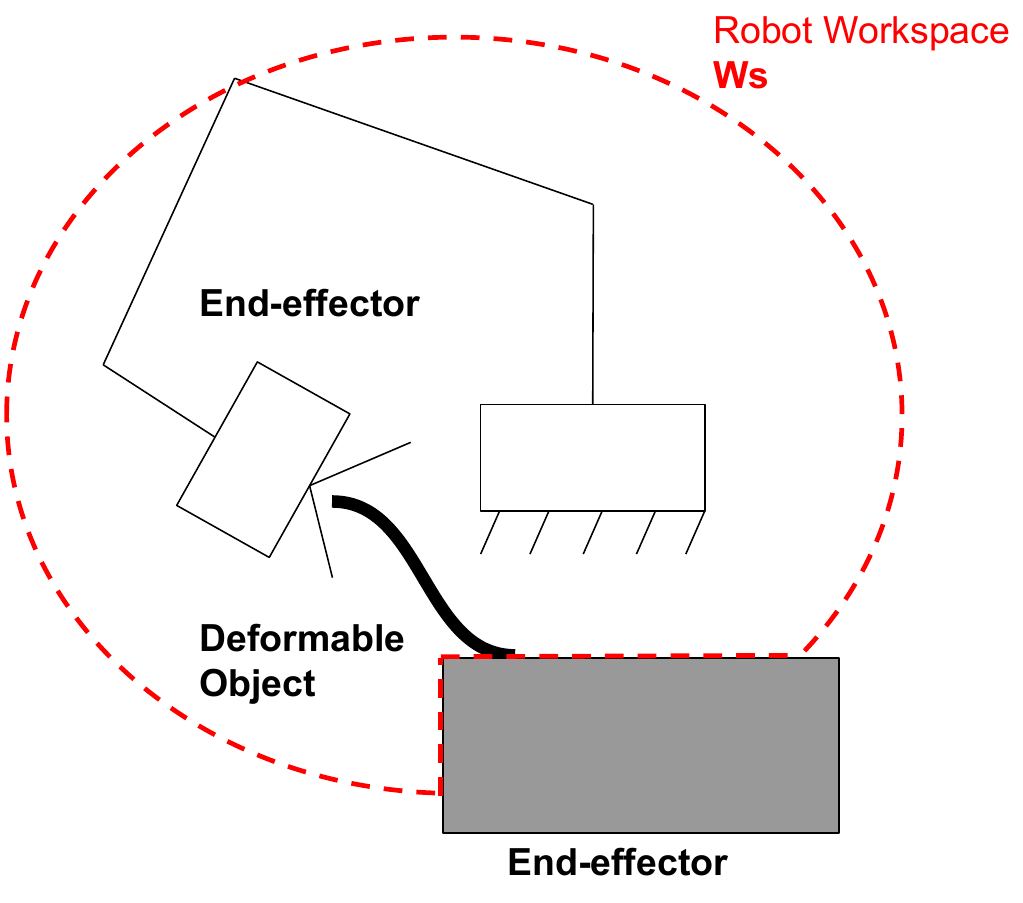}
        \caption{schematic representation of a robotic end-effector's operational workspace Ws}
        \label{robot_ws}
    \end{figure}

\subsection{Numerical resolution}

\noindent The Cosserat forward kinematic model is useful to predict the deformations of the object but for a motion planning application, we need to be able to generate the command to send to the robot in order to control the shape of the manipulated object. However, the Cosserat forward kinematic model cannot be easily inverted so we cannot use an analytical solution. Instead, we numerically solve the problem with an algorithm based on the shooting method which is a classical approach for solving boundary value problems. The Cosserat equations are integrated along the rod until boundary conditions that satisfy the constraints are found. On top of this, there is another optimization loop that ensures the optimality of the solution where the optimization variables are the unset initial conditions of the problem. When integrating the model equation with the optimized variables, we obtain an optimal configuration of the rod with respect to the cost function (see Figure \ref{opt_proc}).

    \begin{figure}[!t]
        \centering
        \includegraphics[scale=0.29]{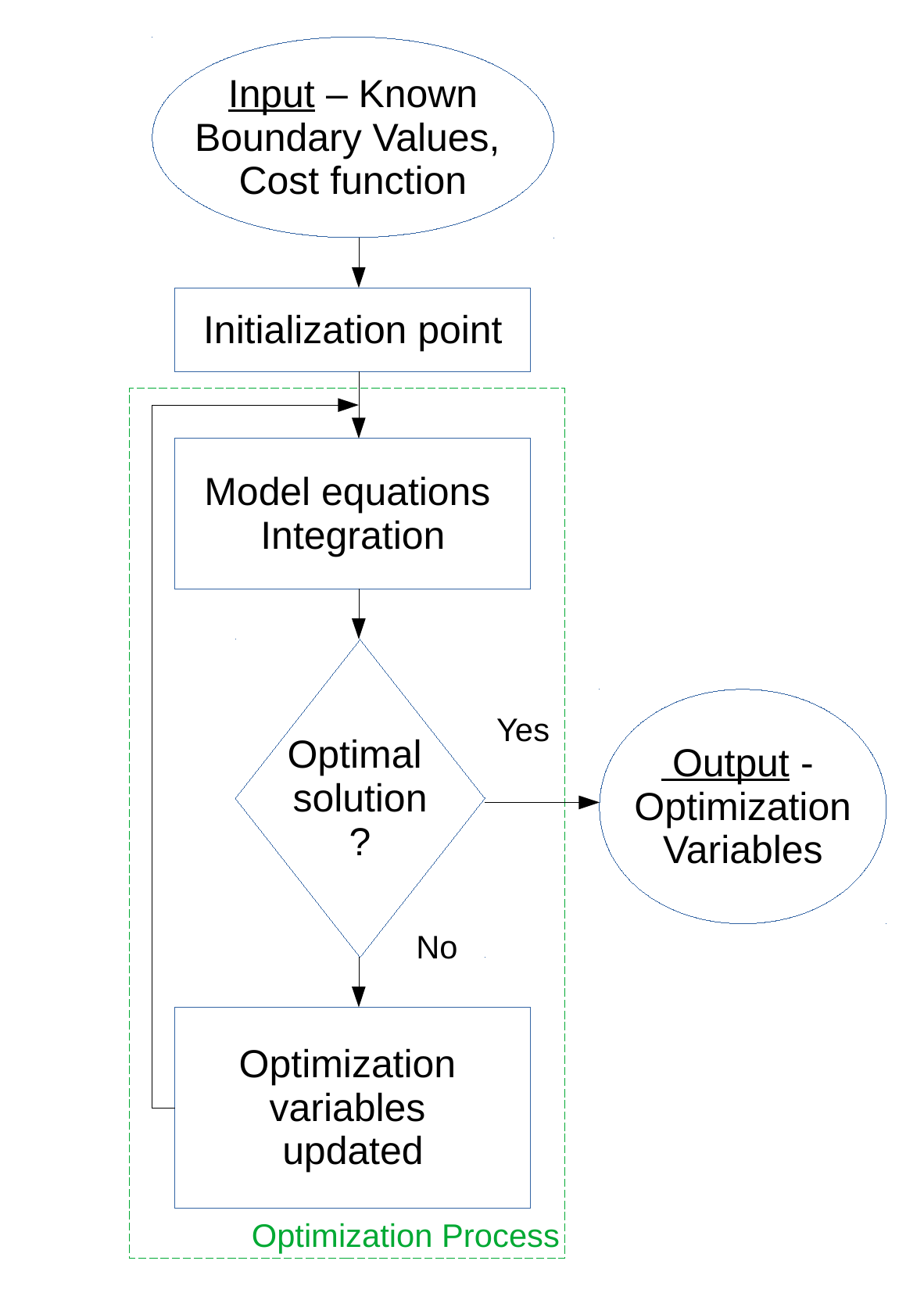}
        \caption{Optimization process}
        \label{opt_proc}
    \end{figure}

\subsection{Optimal trajectory generation}

\noindent Interior-point methods are used for the optimization therefore we need to provide the algorithm an initial guess "close enough" to the solution to start the optimization. 
Since this problem is a highly nonlinear boundary value problem, providing a good initial guess is a key problem as the algorithm can fail to avoid local minima or to find a solution otherwise. 

To ensure the algorithm does not fall in local minima, we introduce $n_{inter}$ intermediate object configurations between the initial known and final configurations. To do this, we define $p^{init}$ as the projection of $p^{obj}$ on the initial shape and discretize the positions between $p^{obj}$ and $p^{init}$ into $n_{inter}$ sets of intermediate objective points configurations. We then iterate and solve the optimization problem at each step by providing the results from the previous step as the initial guess for the next step.

As a by-product, we also obtain a set of intermediate optimal configurations that serve as a trajectory for the object to deform from its initial to the desired configuration. In \cite{campisano2021closed}, the trajectory is defined by discretizing the end-effector's position between the initial and final configurations but the state of the object is not known during the trajectory, the feasibility of the path is therefore not guaranteed. Another advantage is that for real applications the viscosity of the object is often not negligible, each configuration hence depends on the path taken and there is benefit in defining the intermediate steps and controlling the shape of the object during the manipulation.
The intermediate set of objective positions is defined as: 

\begin{equation}
p^{i} = p^{init} + (i-1)\frac{p^{obj}-p^{init}}{n_{inter}}
\end{equation}

Where $i = 1...n_{inter}$, see Figure \ref{trajectory_generation} for an illustration.

We suppose that the contact between the robot and the object remains steady during the manipulation, therefore the configuration of the end-effector corresponds to the configuration of the contact point.
From the output of the algorithm, we can subsequently deduce the position and orientation of the end effector to deform the object into this optimal configuration.

    \begin{figure}[!t]
        \centering
        \includegraphics[width=0.5\textwidth]{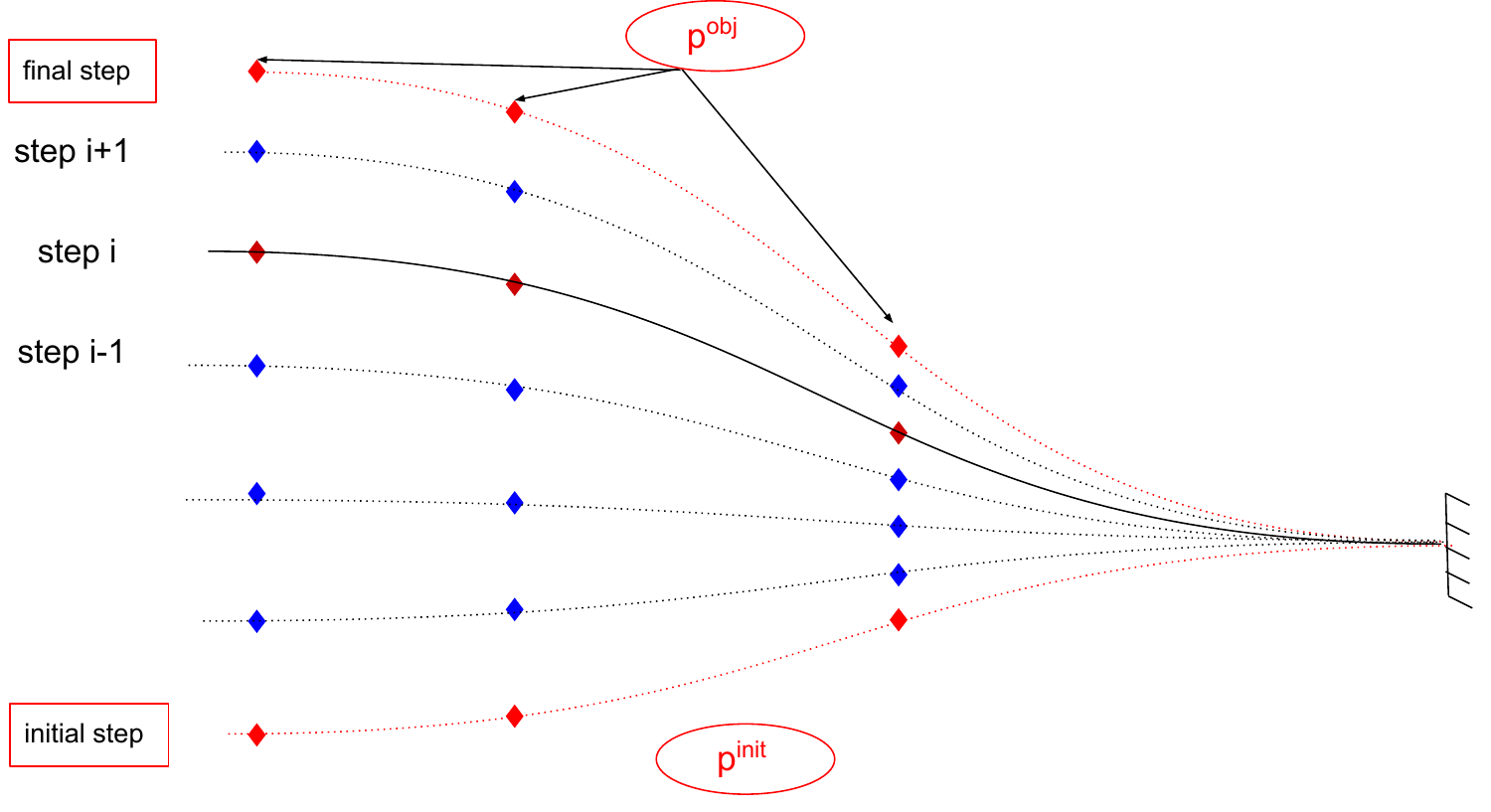}
        \caption{Set of $n_{inter}$ successive optimal configurations from which the end-effector trajectory is derived}
        \label{trajectory_generation}
    \end{figure}

Since we use a static Cosserat model, the end-effector velocity has to be low enough to limit the dynamic effects.

The process can be summarized as (see Figure \ref{trajectory_schema}): 
\begin{itemize}
    \item Define $p^{obj}$;
    \item Input $p^{obj}$ and $p^{init}$;
    \item Define the intermediate objectives between $p^{obj}$ and $p^{init}$;
    \item Solve the optimization process at each step, providing the input for the next step;
    \item Determine the end-effector commands from the intermediate and final configurations of the object.
\end{itemize}

    \begin{figure*}[]
        \centering
        \includegraphics[width = 0.95\textwidth]{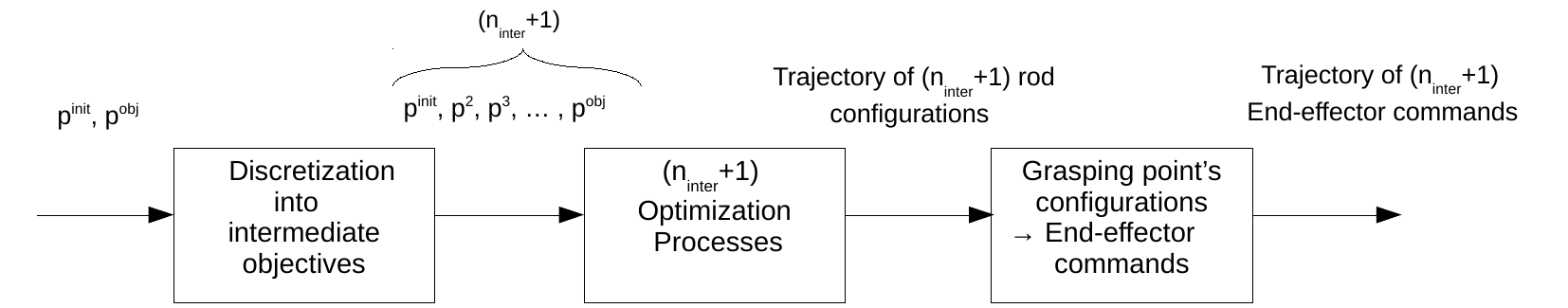}
        \caption{Successive steps to acquire a trajectory from known initial and desired configurations}
        \label{trajectory_schema}
    \end{figure*}

\section{Experimental Validation}
\subsubsection{Experimental Setup}

\begin{figure}[htb]
    \centering
    \includegraphics[scale=0.25]{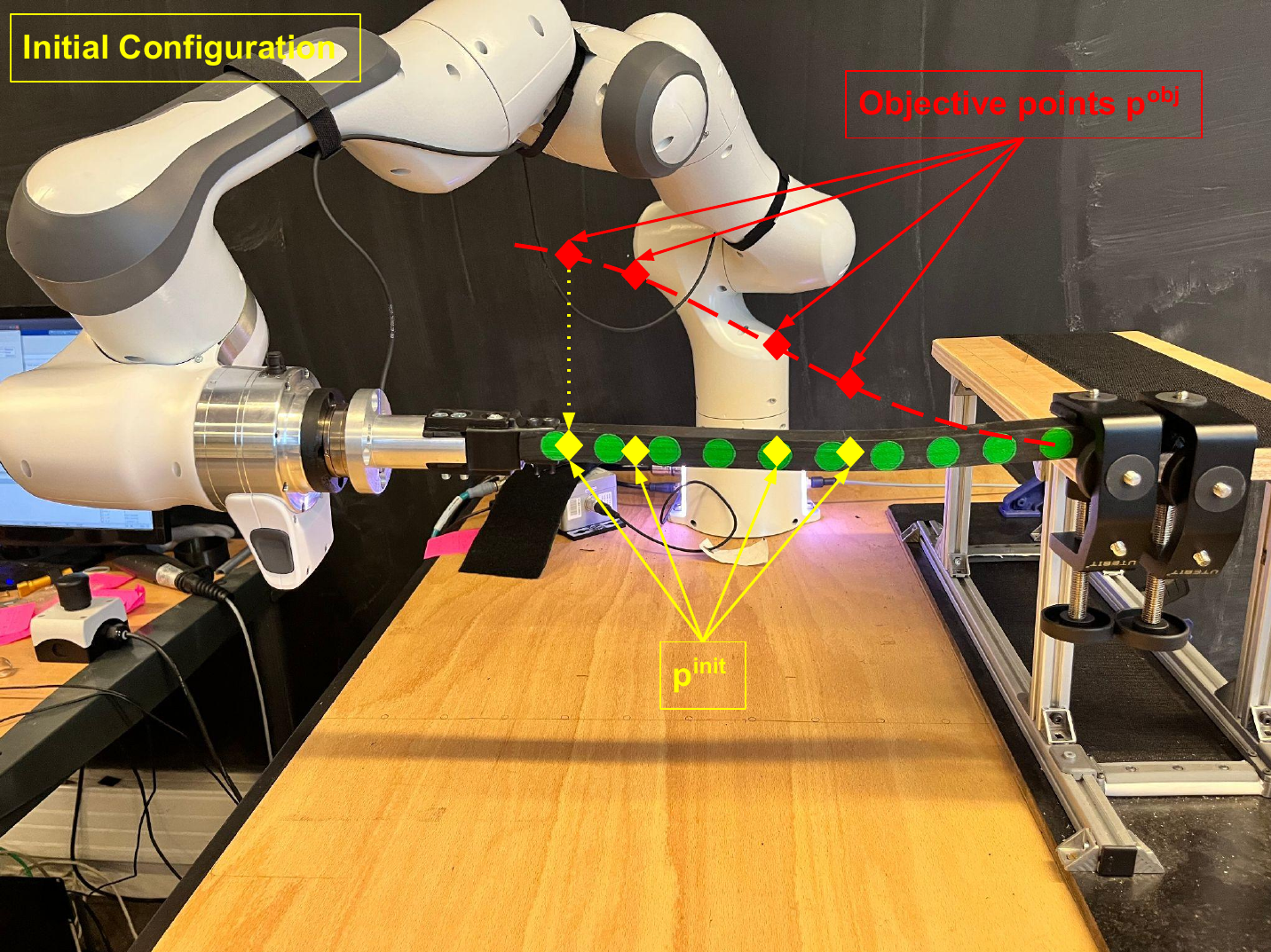}\\
\label{error_3d_multiple_exp}
    \vspace{0.1cm}
    \includegraphics[scale=0.25]{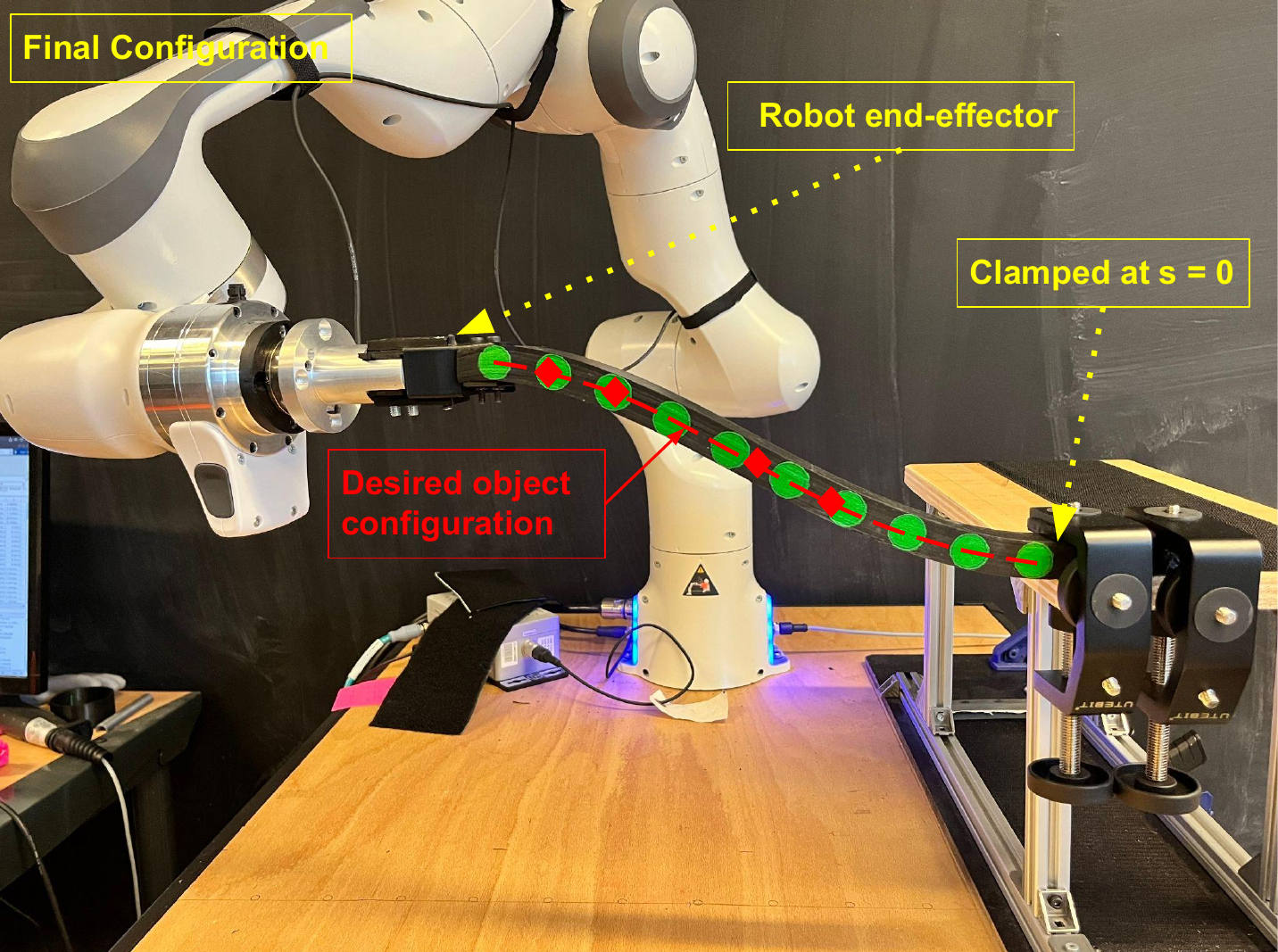}
    \caption{Experimental setup at the initial and final configurations.}
    \label{photo_setup_fonctionne}
\end{figure}

%\begin{figure}[]
%    \centering
%    \subfloat{\includegraphics[scale=0.28]{images/exp_1_a.pdf}}
%    \hspace{0.1cm}
%    \subfloat{\includegraphics[scale=0.28]{images/exp_1_b.pdf}}
%    \caption{Experimental setup at the initial and final configurations.}
%    \label{photo_setup_fonctionne}
%\end{figure}

\noindent To validate the described approach, we propose the following experiment: a robotic arm deforms a linear deformable object, clamped at one end and grasped by the robot at the other end, towards a desired 3D shape. We compare each object pose during the trajectory to the model and then also compare it to a reference trajectory. 
Let us consider the rubber band (see table \ref{tab_param}) clamped so that the initial position $p(0)$ and orientation $O(0)$ are fixed.
We define $p^{obj} = \begin{bmatrix} p^{obj}_{1}  p^{obj}_{2}  p^{obj}_{3}   ...  
 p^{obj}_{n}\end{bmatrix}^{T}$ the objective points that represent the desired shape in $\mathbb{R}^{3}$. The trajectory is generated by discretizing the position of the objective points between the desired and initial configurations into $n_{inter} = 7$ intermediate steps. 

\begin{table}[htb]
    \centering
\begin{tabular}{ |p{3cm}|p{3cm}| }
 %\hline
 %\multicolumn{4}{|c|}{Country List} \\
 \hline
 Parameter & Value \\
 \hline
 E   & $3.2e^{6}$ $N/m^{2}$\\
 $\nu$ &   0.5 \\
  G = $\frac{E}{2(1+\nu)}$ & $1.06e^{6}$ $N/m^{2}$ \\
  $EI_{x}$ & 0.04 $N.m^{2}$\\
  $EI_{y}$ & 0.04 $N.m^{2}$\\
  $GJ$ & 0.028 $N.m^{2}$\\
 $\rho$ & 1392 $kg/m^{3}$ \\
 L &   0.29 $m$  \\
 h & 0.02 $m$  \\
 b & 0.02 $m$ \\
 g & 9.81 $m/s^{2}$  \\
 \hline
\end{tabular}
    \caption{Parameters}
    \label{tab_param}
\end{table}
    
\par
For the experimental setup, we use a Franka Emika Panda seven degrees of freedom arm (see Figure \ref{photo_setup_fonctionne}). The robot is controlled with Moveit and ROS using Franka Emika FCI, developed for research applications. 
The rubber band (see table \ref{tab_param}) is clamped at one end and fixed to the robot thanks to a 3D printed handle that maintains steady the rubber band so that the orientation and position of the end effector correspond to the position and orientation of the extremity of the rubber band (see Figure \ref{photo_setup_fonctionne}).
The setup also includes a 3D Kinect V2 that tracks the position of 10 markers evenly distributed on the rubber band.

\subsubsection{Results}

%\begin{figure*}[t!]
%\centering
%\subfloat[]{\includegraphics[scale=0.055]{images/Figure13.pdf}}
%\subfloat[]{\includegraphics[scale=0.055]{images/Figure14.pdf}}
%\subfloat[]{\includegraphics[scale=0.055]{images/Figure15.pdf}}
%\subfloat[]{\includegraphics[scale=0.055]{images/Figure16.pdf}}
%\subfloat[]{\includegraphics[scale=0.055]{images/Figure17.pdf}}
%\caption{Experimental results for multiple objectives and object shapes}\label{5photos}
%\end{figure*}

\begin{figure*}[t!]
\centering
\includegraphics[width=0.93\textwidth]{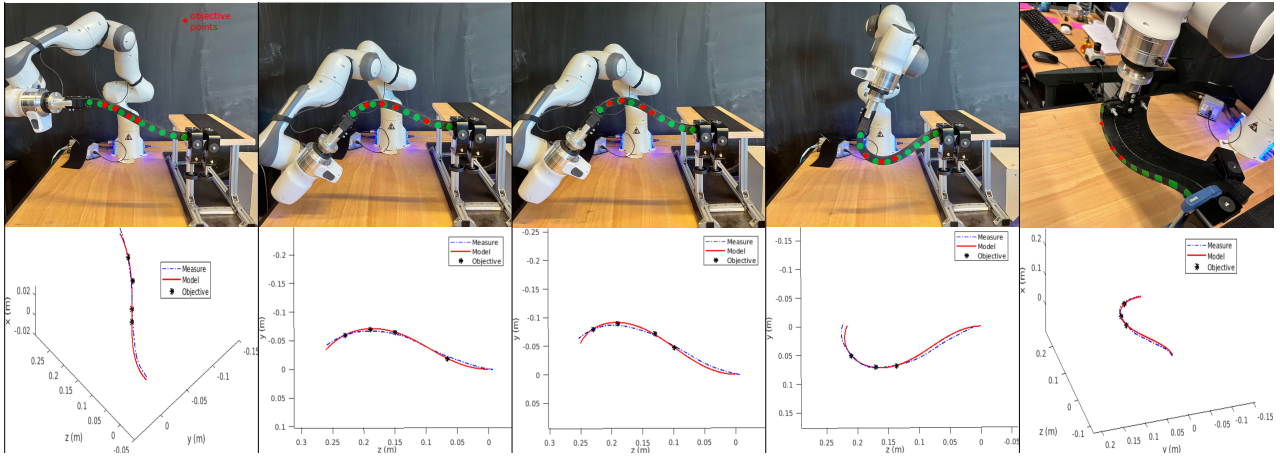}
\caption{Experimental results for multiple objectives and object shapes}\label{5photos}
\end{figure*}

\begin{figure}[htb]
    \centering
    \subfloat[3D Error convergence over time for the five experiments]
    {\includegraphics[scale=0.32]{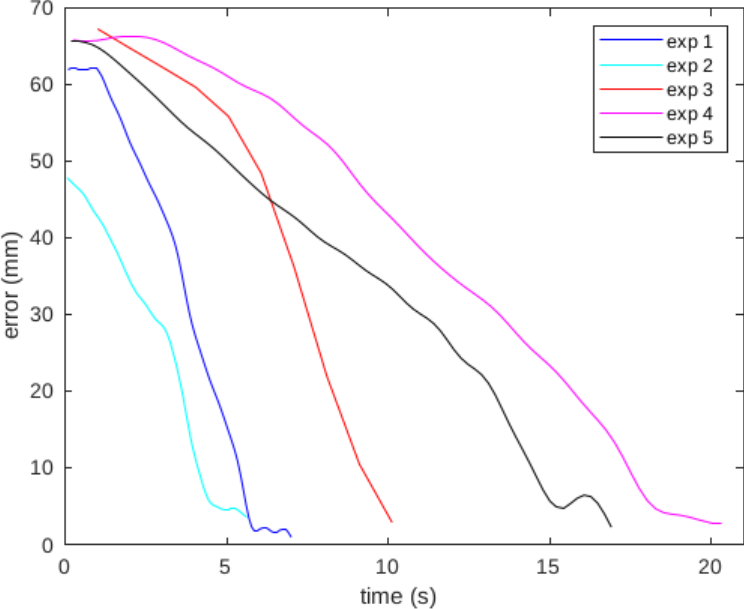}
    \label{error_3d_multiple_exp}}
    \hspace{0.025cm}
    \subfloat[2D Error convergence over time for the five experiments]
    {\includegraphics[scale=0.32]{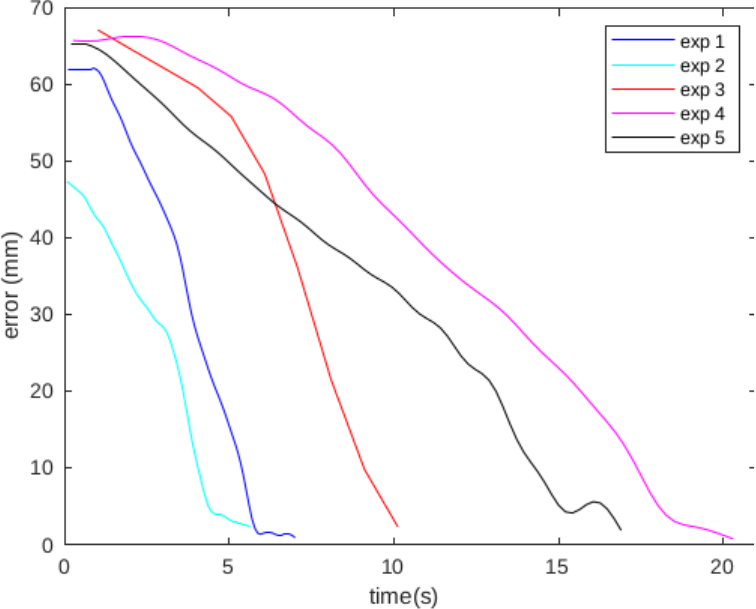}
    \label{error_2d_multiple_exp}}
    \caption{}
    \label{error_over_time}
\end{figure}

\begin{figure}[htb]
    \centering
    \subfloat[Final error variance across multiple experiments with different $p^{obj}$]
    {\includegraphics[scale=0.32]{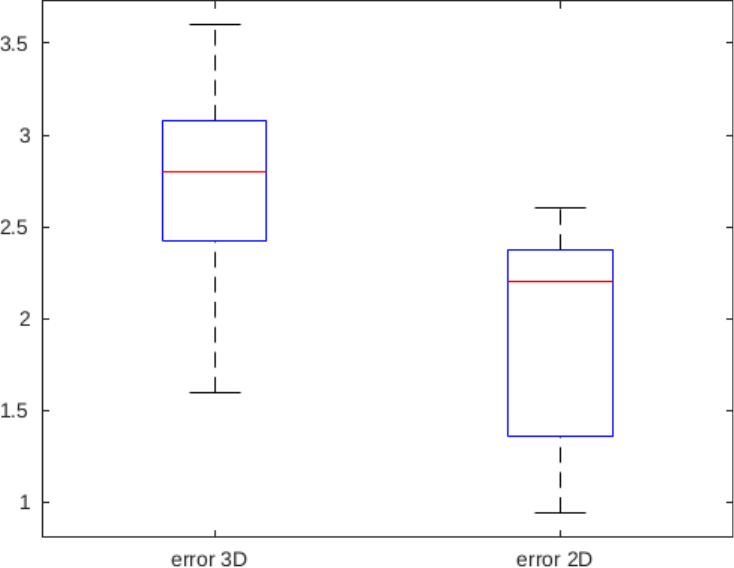}
    \label{boxplotmultipleexpvariance}}
    \hspace{0.025cm}
    \subfloat[Final error variance across the same experiment repeated multiple times]
    {\includegraphics[scale=0.32]{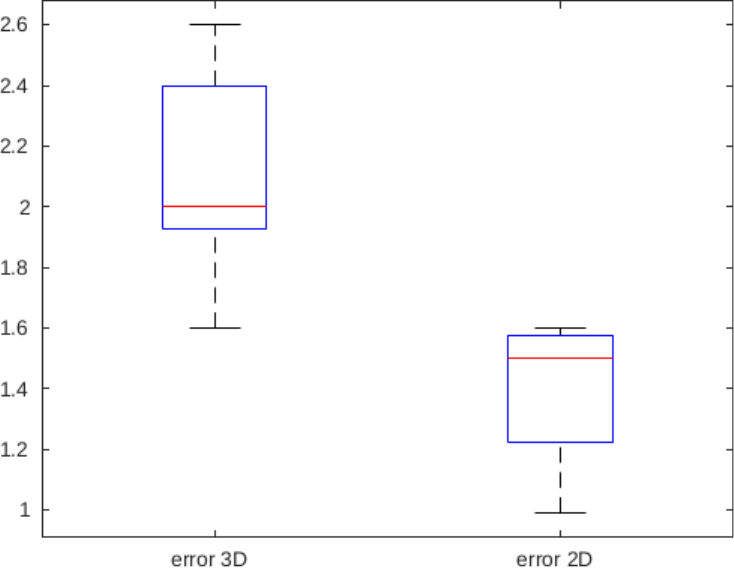}
    \label{boxplot_repeatability_variance}}
    \caption{}
    \label{boxplots}
\end{figure}

%\begin{figure}[]
%    \centering
%    \includegraphics[width=0.45\textwidth]{images/boxplot_multiple_exp_variance_2.pdf}
%    \caption{Final error variance across multiple experiments with different $p^{obj}$}
%    \label{boxplotmultipleexpvariance}
%    \includegraphics[width=0.45\textwidth]{images/boxplot_repeatability_variance_2 .pdf}
%    \caption{Final error variance across the same experiment repeated multiple times}
%    \label{boxplot_repeatability_variance}
%\end{figure}

%%%%%%%%%%%%%%%%%%%%%%%%%%%%%%%%%%%%%%%%%%%%%%%%%%%%%%%%%%%%%%%%
\noindent To validate the approach, we compare the object's configuration to $p^{obj}$ and plot the error over time to show the convergence of the algorithm. The error is the average distance between the measured configuration of the object and each objective point of $p^{obj}$. The robot's end-effector velocity is defined beforehand and is not a source of error as long as the model hypotheses are fulfilled, ie. the object's dynamic effects are negligible. 
To highlight the efficiency of the method over different objects and shapes, we propose five experiments each with different $p^{obj}$ or objects and plot the error over time during the manipulation in Figure \ref{error_over_time}. 
The corresponding experiments are depicted in Figure \ref{5photos}, the upper picture is a photo of the system once the objective is reached and the lower graph represents a comparison between the objective points $p^{obj}$, the model's shape prediction, and the measured state of the object at the final state of the trajectory for the corresponding experiment.
The final step 3D error distribution for the experiments is shown in Figure \ref{boxplots} and ranges from 1.6mm to 3.6mm. At this scale, the camera error is not negligible, especially for the depth measurements that are less precise with the Kinect V2 than in the other directions. We therefore also show the errors in the 2D plane in Figure \ref{boxplots} and \ref{error_2d_multiple_exp}. 
Here we used two different objects, a beam of square section and an elastic band. Any object can be used as long as the Cosserat model's hypotheses are fulfilled. Since the velocity is constant across all  experiments, the robot's travel time is only determined by the path taken. 
\par To show that our approach is general and that the final error is stable over multiple $p^{obj]}$, we compare the error distribution obtained in Figure \ref{boxplotmultipleexpvariance} to the error distribution obtained when repeating the same experiment 8 times with the same inputs (see Figure \ref{boxplot_repeatability_variance}). Repeatability is never ensured when manipulating deformable objects, multiple hypotheses like the elastic behavior and the negligible viscosity of the object are not verified in reality. Hence, the shape of the object depends on the path taken unless the manipulation speed is slow enough to entirely nullify the viscous effects, which is not suited for robotic applications. Also, flaws in the object and errors in the estimation of elastic parameters often lead to unexpected behaviors and biases. These factors can hinder the repeatability of the experiment hence why we repeated the same experiment multiple times to estimate the error variance.
We can see that the distribution for the repeated experiment lies in a similar range as the one with different experiments, which shows the consistency as well as the repeatability to some degree of the method. This must however be confirmed with further experimentation as the sample size is low for both cases.

\section{Discussion and future work}

\noindent In this article, we proposed an approach based on the Cosserat theory to control the shape of linear and planar objects that respect the Cosserat hypothesis. 
We used the framework previously developed for deformable linear robot control and adapted it to deformable object manipulation.
We proposed to reformulate the problem as an optimization problem where the distance between a set of objective points and the 3D curve formed by the object and the elastic energy in the object are minimized. 
To improve the robustness of the algorithm and control the trajectory connecting the initial and final configurations of the object, we discretized the objective points into intermediate objectives in order to define intermediate configurations.
From these previous configurations, we deducted the end-effector positions to deform the object toward the desired shape.
To further improve the results and the final distance error, the next step is to close the loop and add vision and force feedback. There is a trade-off between execution time and algorithm precision and multiple parameters can be influenced in order to reduce it, among them are the number of objective points, the complexity of the desired shape, the number of intermediate objectives, object rigidity, and discretization fineness, weights in the cost function. So far these parameters are tuned towards high precision but adding sensor feedback can help counterbalance the precision loss. Additionally, a spline or curvature-based representation of the object can be used to estimate the load \cite{diezinger20223d} at the limits of the objects.

Two main challenges in this field are handling contacts with the manipulated object as well as manipulating an object with a complex shape. If the contact forces are known (ie. a clip on the object of a known punctual force), then both problems can be solved by having multiple integration sections with different boundary conditions and shapes. The problem is much more difficult however for contacts that are difficult to model (ie. velcro, contact with other soft objects).

\section*{Funding sources}

This work was supported by the SOFTMANBOT project, which received funding from the European Union’s Horizon 2020 research and innovation program under grant agreement No 869855.
\bibliographystyle{IEEEtran}
\bibliography{biblio}
\endgroup
\end{document}